\documentclass[
]{ceurart}

\usepackage{listings}
\usepackage{glossaries}
\usepackage[nodayofweek]{datetime}
\usepackage{cleveref}
\usepackage{amsmath}
\usepackage{amssymb}
\usepackage{tabularx}
\usepackage{listings}
\usepackage{xcolor}

\newdateformat{mydate}{\THEYEAR-\twodigit{\THEMONTH}-\twodigit{\THEDAY}}
\newcommand{\footnoteurl}[1]{\footnote{\url{#1}, last accessed \mydate\today}}
\newcommand{\promptex}{\hspace*{6mm} Example: }
\newtheorem{prompt}{Prompt}
\Crefname{prompt}{Prompt}{Prompts}
\Crefname{lst}{Listing}{Listings}

\colorlet{punct}{red!60!black}
\definecolor{background}{HTML}{EEEEEE}
\definecolor{delim}{RGB}{20,105,176}
\colorlet{numb}{magenta!60!black}

\lstdefinelanguage{json}{
    basicstyle=\ttfamily\footnotesize,
    numbers=left,
    numberstyle=\scriptsize,
    stepnumber=1,
    numbersep=8pt,
    showstringspaces=false,
    breaklines=true,
    frame=lines,
    literate=
     *{0}{{{\color{numb}0}}}{1}
      {1}{{{\color{numb}1}}}{1}
      {2}{{{\color{numb}2}}}{1}
      {3}{{{\color{numb}3}}}{1}
      {4}{{{\color{numb}4}}}{1}
      {5}{{{\color{numb}5}}}{1}
      {6}{{{\color{numb}6}}}{1}
      {7}{{{\color{numb}7}}}{1}
      {8}{{{\color{numb}8}}}{1}
      {9}{{{\color{numb}9}}}{1}
      {:}{{{\color{punct}{:}}}}{1}
      {,}{{{\color{punct}{,}}}}{1}
      {\{}{{{\color{delim}{\{}}}}{1}
      {\}}{{{\color{delim}{\}}}}}{1}
      {[}{{{\color{delim}{[}}}}{1}
      {]}{{{\color{delim}{]}}}}{1},
}
\newacronym{ai}{AI}{Artificial Intelligence}
\newacronym{bfo}{BFO}{Basic Formal Ontology}
\newacronym{dl}{DL}{Description Logic}
\newacronym{gpt}{GPT}{Generative Pretrained Transformer}
\newacronym{kbs}{KBS}{Knowledge-Based System}
\newacronym{kg}{KG}{Knowledge Graph}
\newacronym{llm}{LLM}{Large Language Model}
\newacronym{lm}{LM}{Language Model}
\newacronym{nlp}{NLP}{Natural Language Processing}
\newacronym{owl}{OWL}{Web Ontology Language}
\newacronym{rdf}{RDF}{Resource Description Framework}
\newacronym{swt}{SWT}{Semantic Web Technologies}
\newacronym{tbox}{TBox}{Terminological Component}

\begin{document}

\copyrightyear{2023}
\copyrightclause{Copyright for this paper by its authors.
  Use permitted under Creative Commons License Attribution 4.0
  International (CC BY 4.0).}

\conference{ISWC 2023 Posters and Demos: 22nd International Semantic Web Conference, November  6--10, 2023, Athens, Greece}

\title{Exploring Large Language Models as a Source of Common-Sense Knowledge for Robots}

\author[1]{Felix Ocker}[%
email=felix.ocker@honda-ri.de,
]
\cormark[1]

\author[1]{J{\"o}rg Deigm{\"o}ller}[%
email=joerg.deigmoeller@honda-ri.de,
]

\author[1]{Julian Eggert}[%
email=julian.eggert@honda-ri.de,
]

\address[1]{Honda Research Institute Europe, Carl-Legien-Str. 30, 63073 Offenbach am Main, Germany}

\cortext[1]{Corresponding author.}

\begin{abstract}
Service robots need common-sense knowledge to help humans in everyday situations as it enables them to understand the context of their actions.
However, approaches that use ontologies face a challenge because common-sense knowledge is often implicit, i.e., it is obvious to humans but not explicitly stated.
This paper investigates if \glspl{llm} can fill this gap.
Our experiments reveal limited effectiveness in the selective extraction of contextual action knowledge, suggesting that \glspl{llm} may not be sufficient on their own.
However, the large-scale extraction of general, actionable knowledge shows potential, indicating that \glspl{llm} can be a suitable tool for efficiently creating ontologies for robots.
This paper shows that the technique used for knowledge extraction can be applied to populate a minimalist ontology, showcasing the potential of \glspl{llm} in synergy with formal knowledge representation.
\end{abstract}


\maketitle

\section{Introduction}
To assist humans effectively in everyday life, service robots need a sound foundation of common-sense knowledge to guide their actions.
While there has been significant research on \glspl{kbs} for robotics \cite{olivares2019review}, creating the underlying knowledge bases remains a considerable challenge \cite{gupta2004common}.
This is due to the fact that common-sense knowledge is inherently implicit, i.e., it is naturally understood by humans but remains mostly unexpressed.
However, the recent advances in \glspl{llm} have opened up a potential source of such knowledge.
This paper explores state-of-the-art \glspl{llm} as a source of common-sense knowledge for robots, and investigates if they could even replace \glspl{kbs}.

\section{Related Work}
\label{sec:sota}

Common-sense knowledge for robots involves actions necessary to reach a desired state, defined by an action verb, an actor, an object, and a tool, forming an action pattern \cite{eggert2020action}.
Existing community efforts, e.g., ConceptNet \cite{speer2017conceptnet}, offer valuable information but face scalability challenges and heterogeneity due to the variety of contributors.
\glspl{llm} such as OpenAI's ChatGPT and BigScience's BLOOMZ seem like a promising alternative due to the large amounts of common-sense knowledge they incorporate.
As the usability of \glspl{llm} increases, knowledge extraction methods have evolved from demasking \cite{losing2021extraction} to zero-shot prompts \cite{caufield2023structured}.

The application of \glspl{llm} to robotics leads to the emergence of embodied \gls{ai}.
However, success rates of such systems are still limited, for instance, up to 85~\% for tidying up \cite{wu2023tidybot} and 48~\% for task planning \cite{singh2023progprompt}, posing challenges for robotics applications.
Until the performance of \glspl{llm} improves further, a synergistic use with \glspl{kbs} seems reasonable.

\section{Extracting Action Patterns from Large Language Models}

\subsection{Formalizing Action Patterns}
\label{subsec:formal-action-patterns}

This work focuses on common-sense knowledge in the form of action patterns.
An action pattern $AP$ is represented by an action $a$, which is executed by a set of \textit{Agents} $\mathcal{A}$ using a set of \textit{Tools} $\mathcal{T}$ to modify a set of \textit{Objects} $\mathcal{O}$, cp. \Cref{eq:ap}.
These patterns capture various scenarios a robot might encounter and how it can act to change these situations.
This knowledge allows the robot to infer appropriate actions to fulfill its goals, such as serving bread.
Note that the knowledge should be grounded in real-world settings and continuously tested, possibly resulting in the robot confirming, discarding, or refining its knowledge.
Note that the states of all elements before and after the action can be described via a set of attributes $\mathcal{S}_{before}$ and $\mathcal{S}_{after}$, respectively.
In addition, the spatial relations $\mathcal{R}_{spatial}$ between the elements involved may change.

\begin{equation}
\label{eq:ap}
AP = (a, \mathcal{A}, \mathcal{O}, \mathcal{T}, \mathcal{S}_{before}, \mathcal{S}_{after},R_{spatial})
\end{equation}

The set $\mathcal{AP}$ comprises all valid action patterns in the sense that they are meaningful in real-world applications.
For instance, when tasked with baking bread, a robot has to identify suitable tools, as illustrated by the incomplete action pattern in \Cref{eq:ap-concrete}.

\begin{equation}
\label{eq:ap-concrete}
AP_{bake} = (bake, \{robot\}, \{bread\}, \{\textit{\textbf{\$tool}}\}, \{cold\}, \{hot\}, \{outside\})
\end{equation}

Action patterns can be expressed using \gls{owl} 2 \gls{dl}, a subset of predicate logic.
Here, \textit{Actions} are defined by their relations to the notions \textit{Object}, \textit{State}, \textit{Location}, and \textit{Time}, forming action patterns.
These notions, rooted in previous work, cp. \Cref{sec:sota}, can be aligned with top-level ontologies such as Basic Formal Ontology.
An \textit{Object} can also take the role of an \textit{Agent} or a \textit{Tool}, and spatial relations can be expressed using existing ontologies.
\Cref{eq:ap-axiom} shows the \gls{dl} representation of the notions \textit{Action} and \textit{Object}.

\begin{equation}
\label{eq:ap-axiom}
\begin{split}
Action \ \sqsubseteq \quad & \exists \ \geqslant 1.has\_agent.Object \ \sqcap \ \exists \ \geqslant 1.has\_object.Object \ \sqcap \ \exists \ \geqslant 0.has\_tool.Object \ \sqcap \\
& \exists \ = 1.has\_location.Location \ \sqcap \ \exists \ = 1.has\_time.Time \\
Object \ \sqsubseteq \quad & \exists \ \geqslant 1.has\_state.State
\end{split}
\end{equation}

\subsection{Extracting Parts of Action Patterns}

In order to replace \glspl{kbs}, \glspl{llm} would need to be capable of consistently answering queries such as ``Which tool can I use to bake bread?''.
This task can be posed generically using \Cref{prompt:slot-extraction}. Note that specifying \textit{candidates} corresponds to a simple grounding in reality.
Further, we modified \Cref{prompt:slot-extraction} to extract the state of an object before and after a given action has been applied to it.
The prompt can also be adapted to extract details about the spatial relationships between the object and the tool used.

\begin{prompt}
\label{prompt:slot-extraction}
In the following, I will ask you a question. In your response, I want you to answer with nothing but a list of suitable comma-separated words sorted by relevance.
Which tool can I use to \textbf{\$action} \textbf{\$object}?
Choose only from the following candidates: \textbf{\$candidates}. \\
\promptex ... Which tool can I use to bake bread? Choose only from the following candidates: bowl, oven, knife, ...
\end{prompt}

\subsection{Extracting Entire Action Patterns}

The large-scale extraction of action patterns to incorporate into a knowledge base can be achieved using \Cref{prompt:ap-extraction}.
Compared to \Cref{prompt:slot-extraction}, this prompt specifies the domain of interest and permits constraints within the action pattern, offering additional flexibility.
The information extracted by \Cref{prompt:ap-extraction} may not be directly applicable, but it is well-suited for populating an ontology with broad common-sense knowledge on a large scale.

\begin{prompt}
\label{prompt:ap-extraction}
Please respond with nothing but lists of the form '(action, agent, object, tool)'. An action pattern is defined by an action, i.e., a verb, an agent who executes the action, an object, which is modified, and optionally a tool. Generate \textbf{\$number} action patterns for the domain of interest '\textbf{\$domain\_of\_interest}'. \\
\promptex ... Generate 100 action patterns for the domain of interest 'kitchen'.
\end{prompt}

\section{Experiments}
\label{sec:experiments}

To evaluate the \glspl{llm}' suitability as a source of common-sense knowledge, we used a ground truth consisting of 97 action patterns.
The data set was created in a study with 20 participants, who completed the action patterns by determining the states of each object before and after the action, identifying the tools used, and outlining the spatial relationships between the objects and tools.
An excerpt of this ground truth for the action ``cut bread'' is shown in \Cref{lst:gt}.

\begin{lstlisting}[language=json,firstnumber=1,caption=Excerpt from the ground truth for action patterns.,label=lst:gt]
{
    "action": "cut",
    "object": "bread",
    "tools": ["knife", "fork", "key"],
    "object_states_before": ["fresh", "uncut", "whole", "unsliced", ...],
    "object_states_after": ["fresh", "cut", "crusty", "sliced", ...],
    "spatial_relations": ["bread near knife", ...]
},
\end{lstlisting}

\subsection{Architecture and Models}

\begin{figure}[ht]
\centering
\includegraphics[width=\textwidth]{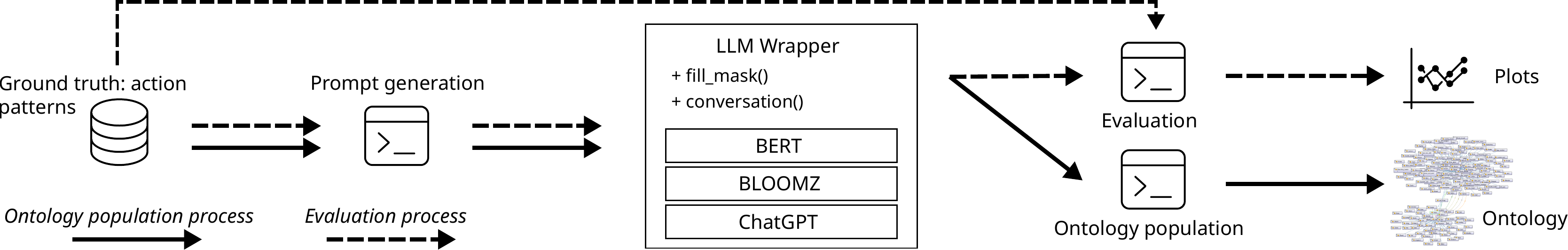}
\caption{Architecture for assessing the \glspl{llm} and populating an ontology.}
\label{fig:architecture}
\end{figure}

We conducted preliminary experiments for choosing suitable \glspl{llm}, including ones with Vicu\~{n}a's 7B and 13B models and OpenAssistant. Despite performing well in chat-like scenarios, these models had difficulty returning structured results. OpenAI's models appeared particularly promising, so we included them along with BLOOMZ as an open-source alternative and BERT as a baseline. For BERT, we rephrased the prompts as demasking tasks.
We accessed gpt-3.5-turbo and gpt-4 via the OpenAI API and BLOOMZ via the Hugging Face API, and deployed BERT locally using a Hugging Face transformers pipeline.
While we experimented with various temperatures, we ended up using a temperature of $0$ to ensure reproducability.
Providing candidates leads to significantly better results when comparing the responses to the ground truth.
However, it is valid to assume that candidates are available, as this corresponds to a grounding in the robot's environment, which can be achieved using an appropriate perception module.
The architecture we developed for the evaluation is shown in \Cref{fig:architecture}.
We also adapted this architecture to populate an ontology revolving around the notions described in \Cref{subsec:formal-action-patterns} with the action patterns extracted to demonstrate the use of the information extracted, see \Cref{fig:architecture} bottom right.

\subsection{Results}

\begin{figure}[ht]
\centering
\includegraphics[width=\textwidth]{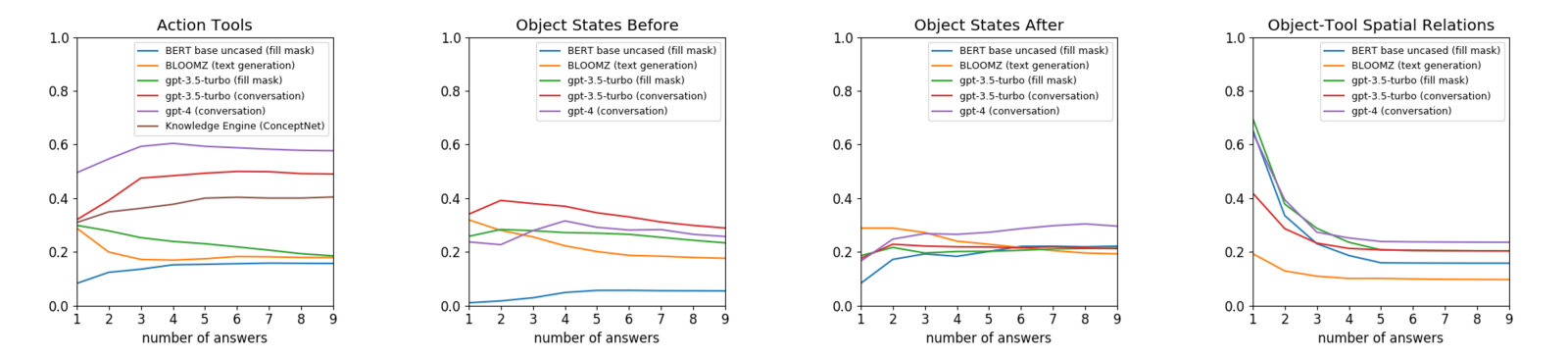}
\caption{F1@n-scores, with n on the x-axis, for the extraction tasks compared to the ground truth ($temperature=0$, candidates provided).}
\label{fig:f1scores}
\end{figure}

\Cref{fig:f1scores} summarizes the results.
It shows F1@n-scores, i.e., the harmonic mean of precision and recall considering the first n elements from the ground truth.
The left plot shows that the OpenAI models outperform others, including a ConceptNet-based baseline, in extracting tools $\mathcal{T}$.
For extracting spatial relations $\mathcal{R_{spatial}}$, demasking was particularly effective.
Extracting object states before $\mathcal{S_{before}}$ and after $\mathcal{S_{after}}$ the action yielded worse results.
Changes in model temperature had little impact.
These results are in line with success rates in \gls{llm} applications \cite{wu2023tidybot, singh2023progprompt}.
To also evaluate the general usefulness of \glspl{llm} for common-sense knowledge extraction, we extracted 100 action patterns for the ``kitchen'' domain.
Upon manual review, all 100 action patterns retrieved were found to be valid.

\section{Conclusion and Outlook}

Our experiments indicate that \glspl{llm} are a scalable source for general common-sense knowledge in the form of action patterns, which is a valuable basis for an ontology usable by robots.
However, \glspl{llm} are still insufficiently reliable for providing actionable knowledge for robotics applications by themselves.
Thus, we suggest pursuing the combination of \glspl{llm} and knowledge graphs for robotics, especially in the form of an integrated pipeline for ontology population from \glspl{llm}, and validation and reasoning using symbolic \gls{ai}.
Future work should also address the validation of newly extracted information using the existing knowledge graph in the spirit of Pan et al. \cite{pan2023unifying}, and the continuous evolution of the knowledge base using all sources available, ranging from databases created in community efforts to \glspl{llm}.

\paragraph*{Supplementary Material:}
The ground truth dataset, the prompts, and a Python script for creating an \gls{owl} ontology from the action patterns extracted are available via GitHub\footnote{\url{https://github.com/HRI-EU/common_sense_for_robots}}.

\bibliography{refs}

\end{document}